\documentclass[times,twocolumn,final]{elsarticle}
\usepackage{ulem}
\usepackage{float}
\usepackage{subfigure}
\usepackage{algorithm}
\usepackage{algorithmic}
\usepackage{medima}
\usepackage{framed,multirow}
\usepackage{multirow, multicol}
\usepackage{threeparttable}
\usepackage{amssymb}
\usepackage{latexsym}
\usepackage{amsmath}
\usepackage{makecell}
\usepackage{xcolor}
\definecolor{newcolor}{rgb}{.8,.349,.1} 
\usepackage{hyperref}

\usepackage{booktabs}
\usepackage{multirow}

\journal{Medical Image Analysis}

\begin{document}

\verso{Wei \textit{et~al.}}

\begin{frontmatter}
\title{SAM-Swin: SAM-Driven Dual-Swin Transformers with Adaptive Lesion Enhancement for Laryngo-Pharyngeal Tumor Detection}

\author[1]{Jia \snm{Wei}\fnref{equal}}
\author[2]{Yun \snm{Li}\fnref{equal}}
\author[1]{Xiaomao \snm{Fan}\corref{cor1}}
\author[3]{Wenjun \snm{Ma}}
\author[1]{Meiyu \snm{Qiu}}
\author[1]{Hongyu \snm{Chen}}
\author[2]{Wenbin \snm{Lei}\corref{cor1}}

\fntext[equal]{These authors contributed equally to this work.}
\cortext[cor1]{Xiaomao Fan and Wenbin Lei are co-corresponding authors.}

\address[1]{College of Big Data and Internet, Shenzhen Technology University, Shenzhen, China}
\address[2]{The First Affiliated Hospital, Sun Yat-Sen University, Guangzhou, China}
\address[3]{Aberdeen Institute of Data Science and Artificial Intelligence, South China Normal University, Foshan, China}

\begin{abstract}
Laryngo-pharyngeal cancer (LPC) is a highly lethal malignancy in the head and neck region. Recent advancements in tumor detection, particularly through dual-branch network architectures, have significantly improved diagnostic accuracy by integrating global and local feature extraction. However, challenges remain in accurately localizing lesions and fully capitalizing on the complementary nature of features within these branches. To address these issues, we propose SAM-Swin, an innovative SAM-driven Dual-Swin Transformer for laryngo-pharyngeal tumor detection. This model leverages the robust segmentation capabilities of the Segment Anything Model 2 (SAM2) to achieve precise lesion segmentation. Meanwhile, we present a multi-scale lesion-aware enhancement module (MS-LAEM) designed to adaptively enhance the learning of nuanced complementary features across various scales, improving the quality of feature extraction and representation. Furthermore,  we implement a multi-scale class-aware guidance (CAG) loss that delivers multi-scale targeted supervision, thereby enhancing the model's capacity to extract class-specific features. To validate our approach, we compiled three LPC datasets from the First Affiliated Hospital (FAHSYSU), the Sixth Affiliated Hospital (SAHSYSU) of Sun Yat-sen University, and Nanfang Hospital of Southern Medical University (NHSMU). The FAHSYSU dataset is utilized for internal training, while the SAHSYSU and NHSMU datasets serve for external evaluation. Extensive experiments demonstrate that SAM-Swin outperforms state-of-the-art methods, showcasing its potential for advancing LPC detection and improving patient outcomes. The source code of SAM-Swin is available at the URL of \href{https://github.com/VVJia/SAM-Swin}{https://github.com/VVJia/SAM-Swin}.
\end{abstract}


\end{frontmatter}


\section{Introduction}
\label{sec:intro}
Laryngo-pharyngeal cancer (LPC), encompassing both laryngeal and hypopharyngeal cancers, represents 65\% to 70\% of all upper respiratory tract cancers. Reports from 2020 indicate that LPC was responsible for 130,000 deaths \citep{hutcheson2012functional}. Laryngeal cancer, in particular, shows limited responsiveness to radiotherapy and chemotherapy, often necessitating total laryngectomy in advanced stages \citep{jones2016laryngeal, tang2018efficacy}. This procedure significantly diminishes patients' quality of life and contributes to their high mortality rates. In contrast, early-stage LPC can be treated with minimally invasive procedures, achieving a 5-year survival rate of up to 90\%, with minimal impact on the voice function \citep{marioni2006current}. Thus, early and accurate diagnosis is vital for effective treatment, reducing mortality, and improving patient prognosis \citep{nocini2020updates}. Diagnosis of LPC relies on the use of an electronic laryngoscope. Biopsies under an electronic laryngoscope are currently the gold standard for diagnosing LPC \citep{mannelli2016laryngeal}. However, due to the limitations of the endoscopist's skill and experience, there are still instances of missed diagnoses and unnecessary repeated biopsies. Consequently, developing accurate automatic detection methods to assist endoscopists in detecting LPC from laryngoscope images is of great significance.

Recent investigations into deep learning methodologies for tumor detection have categorized these approaches into two primary frameworks: single-branch networks \citep{U-Net, chen2017deeplab, Efficientnet, ViT, UC-DenseNet, MTANet, cai2024towards} and dual-branch networks \citep{wu2021elnet, DLGNet, RadFormer, fu2024ipnet, yang2024cvan}. Single-branch networks, including architectures such as U-Net \citep{U-Net}, DeepLab \citep{chen2017deeplab}, EfficientNet \citep{Efficientnet}, UC-DenseNet \citep{UC-DenseNet}, and MTANet \citep{MTANet}, primarily focus on the extraction of global features from endoscopic images, which are critical for various diagnostic applications. While these networks are proficient at capturing overarching patterns and structures within the images, they often overlook the vital local information present in lesion regions. This local context is essential for accurate tumor detection and characterization, as it can provide nuanced details that are pivotal for distinguishing between benign and malignant tissues.

To address the limitations associated with single-branch networks, researchers have developed dual-branch networks that integrate both global and local feature representations. Architectures such as ELNet \citep{wu2021elnet}, DLGNet \citep{DLGNet}, RadFormer \citep{RadFormer}, IPNet \citep{fu2024ipnet}, and cVAN \citep{yang2024cvan}, exemplify this approach, aiming to enhance the detection capabilities by combining the strengths of both feature types. However, despite the advancements offered by dual-branch networks, significant challenges remain. One of the primary issues is the precise localization of lesions within endoscopic images, where the inherent similarity between the foreground (lesion) and background can complicate the model's ability to differentiate between the two. Additionally, many existing dual-branch networks employ a straightforward concatenation method to merge global and local features before they are processed by fusion modules. This direct integration may fail to exploit the complementary nature of these features, potentially limiting the effectiveness of the feature fusion process. As a consequence, these limitations can adversely impact the overall performance of tumor detection, particularly in complex scenarios such as identifying laryngo-pharyngeal tumors, where accurate localization and characterization are crucial for effective clinical intervention.

To address the aforementioned issues, we introduce SAM-Swin, an innovative Dual-Swin Transformer framework enhanced by the Segment Anything Model 2 (SAM2) for the detection of laryngo-pharyngeal tumors. The SAM-Swin mainly consists of four pivotal components: SAM2-guided lesion location(SAM2-GLLM), whole image branch (WIB), lesion region branch (LRB), and multi-scale lesion-aware enhancement module (MS-LEAM). The SAM-Swin capitalizes on the exceptional object segmentation capabilities of SAM2, facilitating precise delineation of tumor regions. To further augment the model's performance, we incorporate an MS-LAEM, which adaptively strengthens the learning of nuanced complementary features across various scales. Additionally, we propose a multi-scale class-aware guidance (CAG) loss function, designed to provide targeted supervision for the extraction of class-specific features, thereby improving the model's discriminative power across different tumor categories. Extensive experiment results on the First Affiliated Hospital (FAHSYSU), the Sixth Affiliated Hospital (SAHSYSU) of Sun Yat-sen University, and Nanfang Hospital of Southern Medical University (NHSMU) demonstrate that the SAM-Swin is superior to the existing state-of-the-art baselines. To sum up, our contributions can be summarized as follows:

\begin{itemize}
\item[$\bullet$] We propose a novel SAM-driven Dual-Swin transformer network specifically designed for the detection of laryngo-pharyngeal tumors. This innovative approach represents the first application of a dual-branch network architecture tailored for the LPC medical imaging challenge.

\item[$\bullet$] By leveraging the advanced object segmentation capabilities of the SAM2, we pioneerly integrate SAM2 into the SAM-Swin framework, enabling SAM-Swin to achieve highly precise segmentation of the lesion region.

\item[$\bullet$] We propose MS-LAEM designed to adaptively enhance the learning of nuanced complementary features across various scales, improving the quality of feature extraction and representation.

\item[$\bullet$] We introduce the multi-scale CAG loss, a novel approach that employs targeted supervision to facilitate the extraction of class-specific features within the model. This loss function is designed to enhance the model's capacity to differentiate between various tumor categories by providing explicit guidance during training.

\item[$\bullet$] Extensive experiments conducted across three datasets of FAHSYSU, SAHSYSU, and NHSMU demonstrate that SAM-Swin achieves competitive performance, consistently outperforming state-of-the-art counterparts.  
\end{itemize}

The remainder of this paper is structured as follows. Section \ref{sec relwork} provides a concise overview of related work concerning SAM and dual-branch frameworks in medical imaging analysis. Section \ref{sec method} presents the details of the proposed SAM-Swin methodology. Section \ref{sec experiment} outlines the experimental setup, discusses the results and their visualization, and addresses the tuning of hyperparameters. Finally, Section \ref{sec conclusion} offers a summary of this paper.

\section{Related works}
\label{sec relwork}
\subsection{Segment anything model in medical imaging analysis}
The SAM is a foundation model known for its impressive zero-shot segmentation performance across diverse natural image datasets \citep{sam}. Building upon SAM, the SAM2 was recently introduced, offering improved segmentation accuracy and efficiency compared to its predecessor \citep{sam2}. Given SAM and SAM2's outstanding performance in natural images, several studies have explored their applicability in medical imaging \citep{mazurowski2023segment, huang2024segment}.

Experimental results in these studies show that while SAM performs well on specific medical tasks, particularly those involving well-defined anatomical structures, it struggles in more complex or ill-defined cases. Furthermore, it has been observed that fine-tuning SAM for domain-specific medical tasks can significantly enhance its segmentation capabilities. Consequently, the potential of SAM for medical image analysis has sparked interest in various fine-tuning strategies tailored for medical applications. 

Fine-tuning methods can generally be divided into two categories: parameter-efficient fine-tuning (PEFT) and non-PEFT strategies. PEFT approaches, such as LoRA \citep{lora}, have been adopted to reduce the number of trainable parameters while maintaining strong performance \citep{samed, chen2023sam, gong20243dsam, chen2024ma}. However, non-PEFT strategies, which involve fine-tuning more extensive portions of the model, have typically shown better performance. For instance, studies such as \citep{ma2024segment}, \citep{yan2024biomedical}, \citep{zhu2024medical}, and \citep{medsam2} applied non-PEFT strategies, fine-tuning either key modules like the Image Encoder and Mask Decoder, or, in some cases, the entire model. While this requires more computational resources, it often leads to greater domain-specific adaptability and improved segmentation accuracy. Despite the success of fine-tuning strategies across various medical applications, SAM2 has yet to be specifically adapted for laryngo-pharyngeal tumor segmentation. Given the performance advantages of non-PEFT approaches, we chose to apply a non-PEFT fine-tuning strategy for SAM2, aiming to achieve better domain-specific adaptability and enhanced segmentation accuracy.

\subsection{Dual-branch frameworks in medical imaging analysis}
Dual-branch frameworks are designed to integrate both global and local features, significantly enhancing lesion classification accuracy across various medical domains. For instance, Wu et al. \citep{wu2021elnet} proposed a dual-stream network that combines global and local contextual information for esophageal lesion classification, with Fast R-CNN \citep{fastrcnn} used to accurately locate lesion regions. Similarly, Basu et al. \citep{RadFormer} utilized activation heatmaps derived from global feature maps to crop salient regions, and then employed a transformer-based architecture to fuse global and local features for Gallbladder Cancer detection. Wang et al. \citep{DLGNet} introduced a dual-branch neural network for intestinal lesion classification, using Mask R-CNN \citep{he2017mask} to detect regions of interest and combine local and global feature representations. Additionally, Fu et al. \citep{fu2024ipnet} proposed a dual-branch network with hierarchical intersection blocks to extract both global and local features for whole-stage colorectal lesion classification. However, despite the success of these methods, several challenges remain: i) The accuracy and precision of lesion localization are critical. Inaccurate or imprecise localization can severely limit the classification performance of subsequent network stages by introducing irrelevant or misleading information. ii) Moreover, simple concatenation is commonly employed to combine global and local features. However, this approach may not fully leverage the complementary aspects of these features, potentially undermining the effectiveness of the fusion process. As a result, these limitations can negatively impact the overall performance of tumor detection, especially in complex situations like laryngo-pharyngeal tumor identification.

\section{Methods}
\label{sec method}
\subsection{Overview}
\begin{figure*}[tb]
    \centering
    \includegraphics[width=\linewidth]{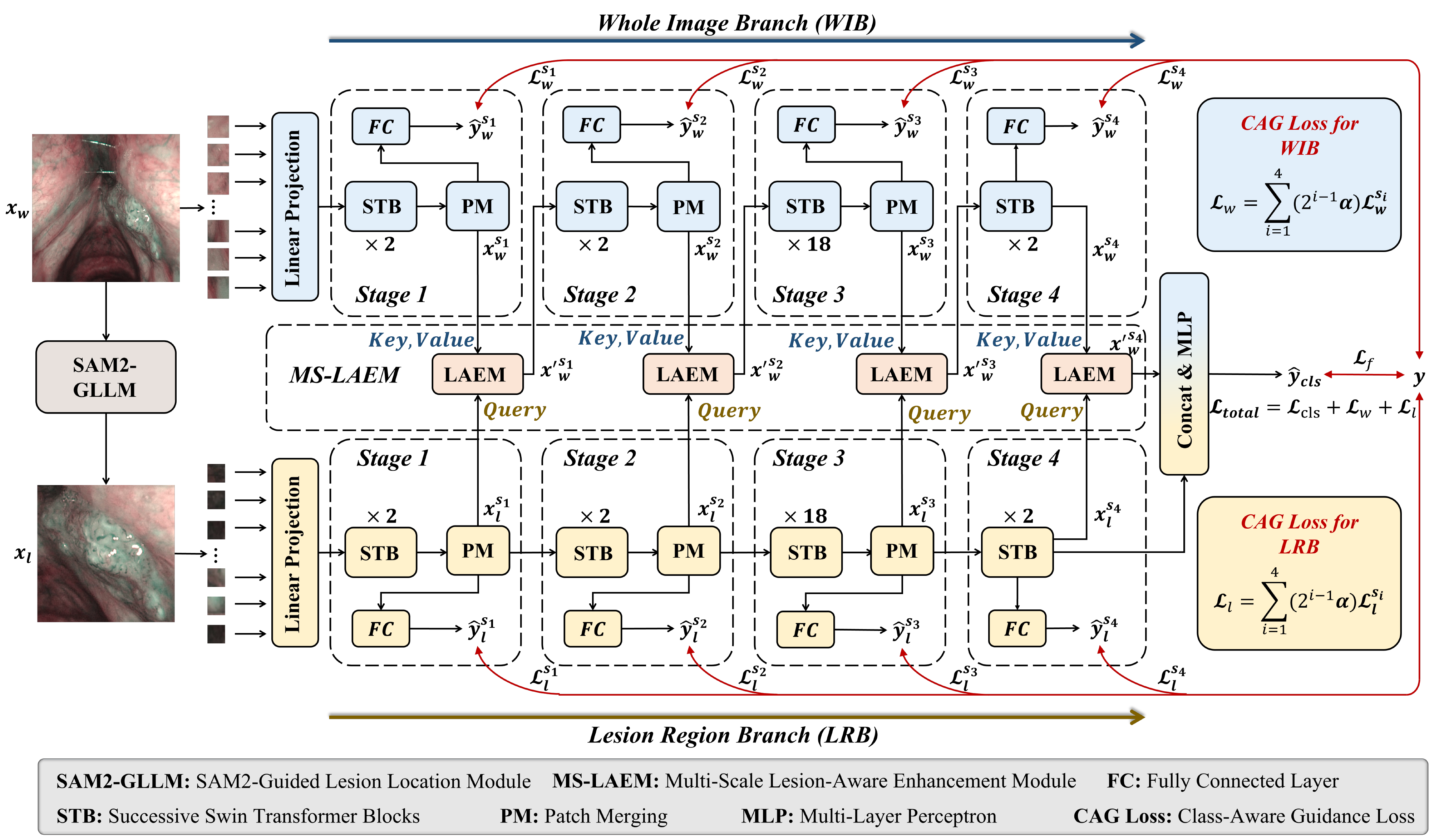}
    \caption{The overall architecture of SAM-Swin. SAM-Swin consists of four key parts: a SAM2-guided lesion location module (SAM2-GLLM), a whole image branch (WIB), a lesion region branch (LRB), and a multi-scale lesion-aware enhancement module (MS-LAEM). 
    }
    \label{F.archite}
\end{figure*}

In this paper, we present SAM-Swin, a novel architecture known as SAM-driven Dual-Swin Transformer with adaptive lesion enhancement, specifically engineered for the detection of laryngo-pharyngeal tumors, as illustrated in Fig. \ref{F.archite}. The SAM-Swin framework comprises four essential components: SAM2-GLLM (discussed in Section \ref{sec:SAM2-GLLM}), WIB (explained in Section \ref{sec:WIB}), LRB (described in Section \ref{sec:LRB}), and MS-LEAM (covered in Section \ref{sec:MS-ALEM}). To further enhance the performance of SAM-Swin, we incorporate multi-scale CAG losses within both the WIB and LRB modules. Additionally, a cross-entropy loss is applied to the final output, as detailed in Section \ref{sec:loss}.

Formally, we define the laryngoscopic image dataset $\mathcal{D} = { (x_w^{(i)}, y^{(i)}) }_{i=1}^N$, which consists of $N$ laryngoscopic images. In this context, $x_w^{(i)}$ represents the holistic image, while $y^{(i)} \in \{0, 1, 2\}$ indicates the classification label, corresponding to normal, benign, and malignant conditions, respectively. The lesion region image $x_l^{(i)}$ is generated through the SAM2-GLLM and shares the identical label $y^{(i)}$ as $x_l^{(i)}$. The SAM-Swin framework employs a multi-task learning approach, integrating multi-scale CAG losses alongside a classification loss. This results in the final outputs denoted $\mathcal{L}_{w}$, $\mathcal{L}_l$, and $\mathcal{L}_{cls}$ for global CAG loss, local CAG loss, and final classification loss, respectively. The overall objective loss for the SAM-Swin framework, denoted as $\mathcal{F}_{SAM-Swin}$, is formulated as follows:
\begin{equation}
    \mathcal{L}_{total} = \mathcal{L}_{cls} + \mathcal{L}_w + \mathcal{L}_l,
\end{equation}
This comprehensive loss function enables the SAM-Swin to effectively balance the contributions from different aspects of the learning process, ensuring robust detection and classification of laryngo-pharyngeal tumors.

\subsection{SAM2-GLLM}
\label{sec:SAM2-GLLM}
\begin{figure}[tb]
    \centering
    \includegraphics[width=\linewidth]{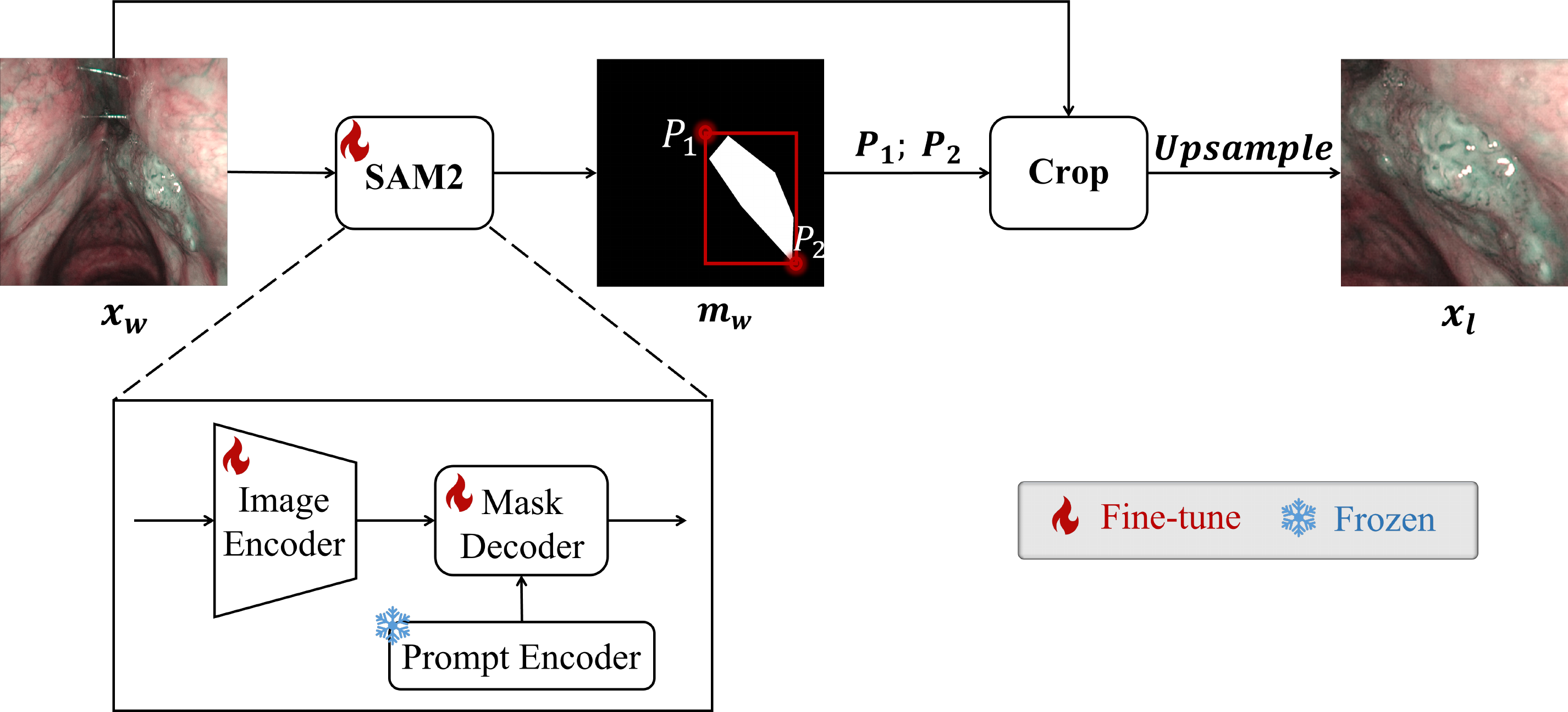}
    \caption{Illustration of the workflow of SAM2-Guided Lesion Location Module (SAM2-GLLM). The whole image $x_w$ is processed by SAM2, generating the corresponding lesion mask $m_w$. Points $P_1$ and $P_2$ are selected based on the foreground, defined as the region where $m_w(x,y)=255$. The lesion region image $x_l$ is then cropped from the $x_w$ using the coordinates of these two points.
    }
    \label{F.sam2_gllm}
\end{figure}

Fig.~\ref{F.sam2_gllm} illustrates the workflow of the SAM2-GLLM framework. In the context of medical imaging, accurately extracting lesion-specific regions from holistic laryngoscopic images is essential for reliable tumor identification. This is typically achieved through object detection or segmentation techniques that localize lesions using bounding boxes or segmentation masks. These localized regions are subsequently cropped for further analysis.

In our approach, we leverage the robust segmentation capabilities of SAM2 to produce high-quality lesion masks. To enhance SAM2's performance in the medical imaging domain, we employ a non-PEFT fine-tuning approach by holistically fine-tuning both the image encoder and mask decoder. This tailored adjustment aims to optimize their functionality for the LPC image segmentation task. Notably, we maintain the prompt encoder in a frozen state to preserve its inherent capacity for effectively processing input prompts. This strategic configuration ensures that SAM2-GLLM is well-equipped for accurate and efficient lesion identification and segmentation.

Specifically, we employ the SAM2, denoted as $\mathcal{F}_{SAM2}$, with the automatic segmentation option (segment-everything) on the whole image $x_w$, yielding the lesion mask $m_w$ as follows:

\begin{equation}
    m_w = \mathcal{F}_{SAM2}(x_w),
\end{equation}

Subsequently, points $P_1$ and $P_2$ are generated based on the foreground defined by the mask:

\begin{align}
    P_1 &= (x_{min}, y_{min}) = (min(x), min(y)), \\
    P_2 &= (x_{max}, y_{max}) = (max(x), max(y)),
\end{align}
\noindent where $(x,y) \in \{ (x,y)\ |\  m_w(x,y)=255 \}$. Finally, the lesion region image $x_l$ is cropped from the whole image $x_w$ using the coordinates of these two points, and then upsampled to match the size of $x_w$:
\begin{equation}
    x_l = \mathcal{F}_{up}(\mathcal{F}_{crop}(x_w, [P_1; P_2])),
\end{equation}

\noindent where $\mathcal{F}_{crop}$ denotes the image cropping function, and $\mathcal{F}_{up}$ represents the upsampling function.

\subsection{WIB}
\label{sec:WIB}
Differed from the work \citep{wei2024sam} with the ResNet \citep{resnet} as the backbone, we adopt a much more powerful Swin Transformer V2 \citep{liu2022swinv2} in this study. This architecture captures long-range dependencies and extracts multi-scale features, thereby enhancing the model's ability to detect and identify lesions. Following the original Swin Transformer V2 architecture, the whole image $x_w \in \mathbb{R}^{H \times W \times 3}$ is first reshaped into a sequence of flattened parches $x_{wp} \in \mathbb{R}^{N \times 3P^2}$, where $H$ and $W$ denote the height and width of the input image, respectively, $P \times P$ is the resolution of each patch, and $N = HW/P^2$ is the total number of patches. In this study, we set the $P$ to 4, following the original Swin Transformer V2 configuration. These patch tokens are then projected into a $C$-dimensional space via a linear projection, resulting in $x_w^{s_0} \in \mathbb{R}^{(\frac{H}{4} \times \frac{W}{4}) \times C}$.

As illustrated in Fig.~\ref{F.archite}, the encoder in the WIB consists of four stages, where each stage comprises Swin Transformer Blocks (STB) followed by a Patch Merging (PM) layer, except for the final stage, which omits the PM layer. The PM layer reduces the resolution of the feature maps by a factor of 2, enabling the encoder to produce hierarchical feature representations. The input tokens $x_w^{s_0}$ are processed through four stages to generate multi-scale features, as follows:
\begin{equation} \label{eq:WIB}
x_w^{s_i} =
\begin{cases} 
\mathcal{F}_w^{s_i}(x_w^{s_{i-1}}), & \text{if } i = 1, \\
\mathcal{F}_w^{s_i}({x'}_w^{s_{i-1}}), & \text{if } i \in \{ 2, 3, 4 \},
\end{cases}
\end{equation}
where $i \in \{  1, 2, 3, 4 \}$ represents the stage index, ${x'}_w^{s_{i-1}}$ is the output from MS-LAEM (described in Section~\ref{sec:MS-ALEM}), and $\mathcal{F}_{s_i}(\cdot)$ denotes the transformation at each stage in the WIB. The resolution of each stage's outputs $x_w^{s_i}$ are $(\frac{H}{8} \times \frac{W}{8}) \times 2C$, $(\frac{H}{16} \times \frac{W}{16}) \times 4C$, $(\frac{H}{32} \times \frac{W}{32}) \times 8C$, and $(\frac{H}{32} \times \frac{W}{32}) \times 8C$, respectively. This hierarchical structure allows the models to capture both fine-grained details and high-level semantic information, which are crucial for accurate lesion classification. Additionally, a fully connected (FC) layer is employed as a projection header following the PM layer at each stage. This layer generates the prediction probability $\hat{y}_w^{s_i}$, which is used for optimizing the CAG loss. 

\subsection{LRB}
\label{sec:LRB}
In this section, we adopt the same architecture as the WIB to ensure consistency in lesion feature extraction. Notably, they do not share the weight parameters, allowing each branch to learn complementary lesion-specific information from different perspectives. Given a lesion region image $x_l$ extracted from a whole image $x_w$ through SAM2-GLLM (described in Section \ref{sec:SAM2-GLLM}), the image is first split into patches. These patches are then reshaped into patch tokens. These path tokens are passed through a linear projection, resulting in $x_l^{s_0} \in \mathbb{R}^{(\frac{H}{4} \times \frac{W}{4}) \times C}$. Further details are provided in Section \ref{sec:WIB}. Subsequently, similar to Eq.~\ref{eq:WIB}, the tokens $x_l^{s_0}$ are processed through four stages, producing multi-scale features as follows:

\begin{equation} \label{eq:LRB}
x_l^{s_i} = \mathcal{F}_l^{s_{i}}(x_l^{s_{i-1}}), 
\end{equation}

\noindent where $i \in \{  1, 2, 3, 4 \}$ denotes the stage index, and $\mathcal{F}_l^{s_i}(\cdot)$ denotes the transformation at each stage in the LRB. Given the rich lesion-specific information present in the lesion region image, the extraction of multi-scale features enables the model to capture more discriminative features, enhancing its ability to distinguish between different categories of lesions. To optimize the CAG loss, an FC layer is utilized as a projection header after the PM layer at each stage. This layer produces the prediction probability $\hat{y}_l^{s_i}$.

\subsection{MS-LAEM}
\label{sec:MS-ALEM}

\begin{figure}[tb]
    \centering
    \includegraphics[width=\linewidth]{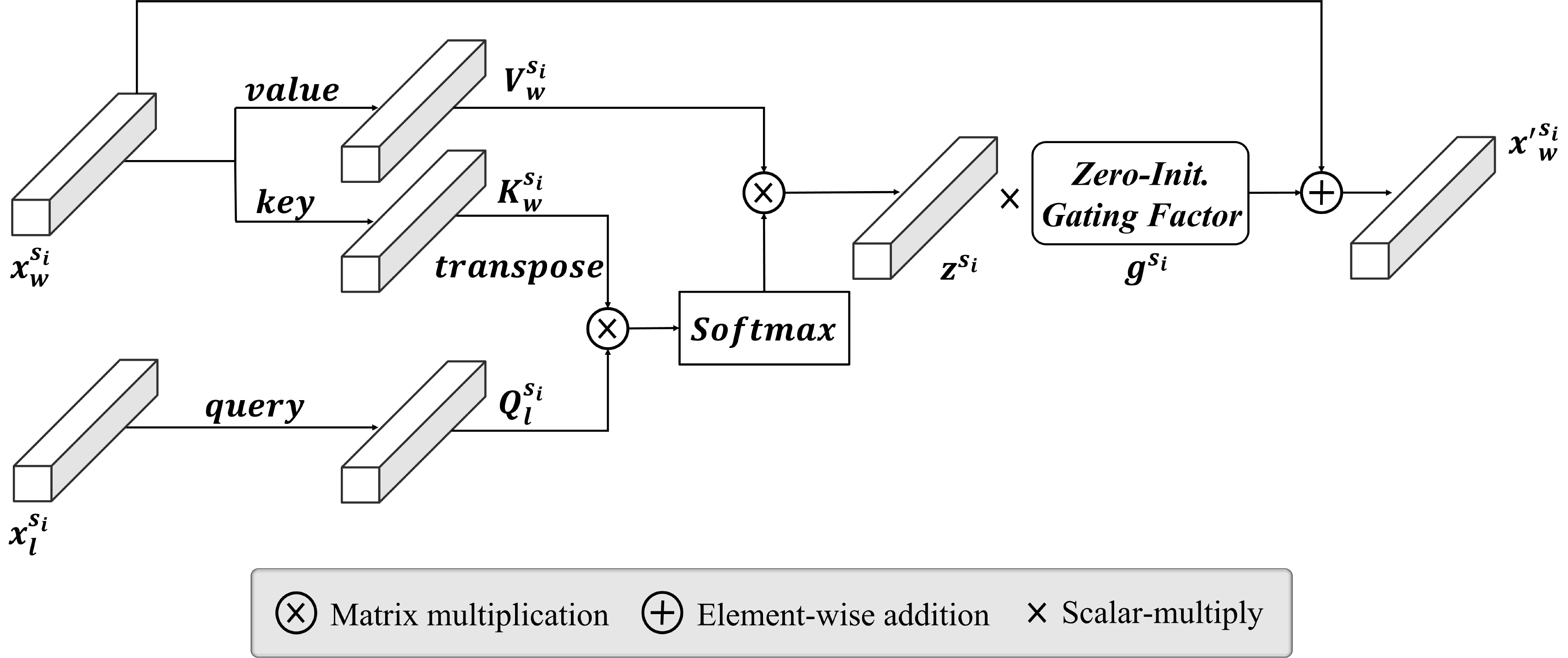}
    \caption{
        The illustration of lesion-aware enhancement module (LAEM). Query tokens are generated from the lesion region tokens, while key and value tokens are produced from the whole image tokens. These query, key, and value tokens are then processed through Multi-Head Attention (MHA) to derive enhanced feature tokens, which contain richer lesion-specific feature representations. Subsequently, the learnable, zero-initialized gating factor is applied to multiply the enhanced feature tokens, adaptively adjusting the importance of the lesion features. Lastly, these enhanced feature tokens are combined with the original whole image tokens to produce the final tokens.
    }
    \label{F.ALEM}
\end{figure}

As shown in Fig.~\ref{F.ALEM}, the whole image tokens $x_w^{s_i}$ (see Eq.~\ref{eq:WIB}) are passed through linear projections to generate the $K_w^{s_i}$ and $V_w^{s_i}$, while the lesion region tokens $x_l^{s_i}$ (see Eq.~\ref{eq:LRB}) are processed to produce $Q_l^{s_i}$. We then perform the Multi-Head Attention (MHA). Specifically, the attention mechanism is applied independently across $h$ heads, and for each head $j$, the attention is computed as:

\begin{equation}
    z_j^{s_i} = Softmax(\frac{Q_{l,j}^{s_i}(K_{l,j}^{s_i})^T}{\sqrt{d_k}})V_{w,j}^{s_i},
\end{equation}

\noindent where $Q_{l,j}^{s_i}$, $K_{l,j}^{s_i}$, and $V_{w,j}^{s_i}$ are the projections for head $j$, $d_k$ is the dimensionality of each head (so that the scaling is done per head), and $Softmax(\cdot)$ is the softmax function. The outputs are obtained by concatenating attention output for all heads,

\begin{equation}
    z^{s_i} = Concat(z_1^{s_i}, z_2^{s_i}, \ldots, z_h^{s_i}).
\end{equation}

\noindent The enhanced feature tokens $z^{s_i}$ represent the lesion-guided features, which probably cause disturbance at the beginning of training. To this end, we adopt a learnable gating factor, denoted as $g^{s_i}$, to adaptively control the importance of $z^{s_i}$. The $g^{s_i}$ is initialized by zero, eliminating the influence of noisy or poorly learned features at the early stages of training. As training progresses, it allows the model to gradually learn the appropriate contribution of lesion-guided features. Thus, the enhanced feature representation is given by:
\begin{equation}
    \hat{z}^{s_i} = g^{s_i} \cdot z^{s_i}.
\end{equation}
Finally, inspired by residual connection mechanism \citep{resnet}, we integrate the enhanced feature tokens $\hat{z}^{s_i}$ back into the whole image tokens $x_w^{s_i}$. This ensures the model retains essential information from the original features while incorporating the refined lesion-aware details for more accurate lesion feature extraction. This integration can be formulated as:
\begin{equation} \label{eq:ms}
    {x'}_w^{s_{i}} = x_w^{s_i} + \hat{z}^{s_i},
\end{equation}
where ${x'}_w^{s_{i}}$ denotes the enhanced tokens for the next stage of processing.

\subsection{Objective loss function} \label{sec:loss}
\subsubsection{Multi-scale CAG loss} \label{sec:cag_loss}
In this section, the stage-specific prediction probabilities $\hat{y}_w^{s_i}$ and $\hat{y}_l^{s_i}$ generated by both WIB and LRB lead to the formulation of the overall multi-scale CAG losses. These losses are defined as follows:
\begin{align}
    \mathcal{L}_w &= \sum_{i=1}^4 (2^{i-1}\alpha) \mathcal{L}_w^{s_i}(\hat{y}_w^{s_i}, y), \\
    \mathcal{L}_l &= \sum_{i=1}^4 (2^{i-1}\alpha) \mathcal{L}_l^{s_i}(\hat{y}_l^{s_i}, y),
\end{align}
\noindent Here, $\mathcal{L}_w^{s_i}(\cdot)$ and $\mathcal{L}_l^{s_i}(\cdot)$ represent the cross-entropy losses at the $i$-th stage on both WIB and LRB, while $\alpha$ serves as a critical trade-off hyperparameter. The exponential weighting $2^{i-1}$ increases with each stage, underscoring the importance of deeper feature representations that capture richer semantic information vital for accurate classification. By allocating greater weight to these deeper features, we enhance the model's ability to focus on significant patterns, thereby mitigating the risk of overfitting to shallower features that may merely capture noise. A detailed discussion on the selection of the hyperparameter $\alpha$ will be presented in Section~\ref{sec:hyper}.

\subsubsection{Classification loss} 
\label{sec:cls_loss}

At the final stage, we obtain the enhanced whole image tokens ${x'}_w^{s_4}$ from MS-LAEM (see Eq.~\ref{eq:ms}), and lesion region tokens $x_l^{s_4}$ from LRB. These two tokens are concatenated and subsequently passed through the MLP layer to produce the final prediction probabilities, denoted as $\hat{y}_{cls}$. The final classification loss $\mathcal{L}_{cls}$ is calculated using cross-entropy as follows:

\begin{equation}
    \mathcal{L}_{cls} = \text{CrossEntropyLoss}(\hat{y}_{cls}, y),
\end{equation}

\section{Experiment}
\label{sec experiment}
\subsection{Experiment settings}
\subsubsection{Datasets}
\begin{table*}[tb]
\centering
\caption{The statistic description of the FAHSYSU, SAHSYSU, and NHSMU datasets.}
\label{dataset_distribution}
\begin{tabular}{@{}lcccccc@{}}
\toprule
\multirow{2}{*}{} & \multicolumn{2}{c}{FAHSYSU (Total=25,256)} & \multicolumn{2}{c}{SAHSYSU (Total=2,788)} & \multicolumn{2}{c}{NHSMU (Total=6,684)} \\ \cmidrule(lr){2-3} \cmidrule(lr){4-5} \cmidrule(lr){6-7}
                  & NBI                 & WLI                 & NBI                & WLI               & NBI               & WLI\\ \midrule
Normal            & 695                 & 7,310               & 9                  & 2,202             & 0                 & 5,772 \\
Benign            & 1,488               & 3,332               & 27                 & 218               & 0                 & 536 \\
Malignant         & 5,954               & 6,477               & 99                 & 233               & 0                 & 376 \\
Total             & 8,137               & 17,119              & 135                & 2,653             & 0                 & 6,684  \\ \bottomrule
\end{tabular}
\end{table*}

All laryngoscopic images in this study are provided by three hospitals, including the First Affiliated Hospital of Sun Yat-sen University (\emph{i.e.}, FAHSYSU), the Sixth Affiliated Hospital of Sun Yat-sen University (\emph{i.e.}, SAHSYSU), and the Nanfang Hospital of Southern Medical University (\emph{i.e.}, NHSMU). 

All laryngoscopic images were collected during routine clinical practice using standard laryngoscopes (ENF-VT2, ENF-VT3, or ENF-V3; Olympus Medical Systems, Tokyo, Japan) and imaging systems (VISERA ELITE OTV-S190, EVIS EXERA III CV-190, Olympus Medical Systems) at an original resolution of $512\times512$ pixels. Our dataset includes laryngoscopic images captured in narrow-band imaging (NBI) mode, compensating for the limitations of white light imaging (WLI) by enhancing the clarity and recognizability of microvasculature \citep{li2021laryngoscopic}. The detailed data descriptions of three datasets are shown in Table~\ref{dataset_distribution}.

\begin{itemize}
    \item \textbf{FAHSYSU:} The FAHSYSU as the internal dataset contains 25,256 images, with 8,137 in NBI mode and 17,119 in WLI mode. It was used for model training, validation, and internal testing.
    \item \textbf{SAHSYSU:} The SAHSYSU as the external dataset contains 2,788 images, with 135 in NBI mode and 2,653 in WLI mode. It was only used for external testing.
    \item \textbf{NHSMU:} The NHSMU as the external dataset contains 6,684 images, only in WLI mode. It was also used for external testing without any data leakage on model training.
\end{itemize}

\subsubsection{Evaluation metrics}
To comprehensively evaluate the performance of our proposed SAM-Swin, we utilized several metrics, including Accuracy, Precision, Recall, and F1-score. These metrics are defined as follows:
\begin{align}
    \text{Accuracy} &= \frac{\text{TP} + \text{TN}}{\text{TP} + \text{TN} + \text{FP} + \text{FN}}, \\
    \text{Precision} &= \frac{\text{TP}}{\text{TP} + \text{FP}}, \\
    \text{Recall} &= \frac{\text{TP}}{\text{TP} + \text{FN}}, \\
    \text{F1-score} &= 2 \times \frac{\text{Precision} \times \text{Recall}}{\text{Precision} + \text{Recall}}.
\end{align}
where the true positive (TP), false positive (FP), false negative (FN), and true negative (TN) values are derived from the confusion matrix. Moreover, Intersection over Union (IoU) and Dice coefficient (Dice) were employed to measure the quality of segmentation produced by SAM2.

\subsubsection{Implementation details} \label{imp_details}
The experiments were performed on a computing server equipped with 4 NVIDIA RTX A6000 GPUs and an AMD EPYC 7763 64-Core Processor running at 1.50 GHz. The implementation was carried out using the PyTorch framework, with Python version 3.10.14, and CUDA version 12.1.

\begin{table}[tb]
    \centering
    \caption{Data distribution for training, validation, and test sets of the FAHSYSU dataset.}
    \label{tab:FAHSYSU_data_distribution}
    \begin{tabular}{lccc}
    \toprule
           & Training & Validation & Test \\ \midrule
    Normal & 5,134 & 571 & 2,301 \\
    Benign & 3,276 & 366 & 1,178 \\
    Malignant & 7,812 & 869 & 3,750 \\
    Total & 16,222 & 1,806 & 7,229 \\ 
    \bottomrule
    \end{tabular}
\end{table}

The test data from the FAHSYSU dataset were randomly selected by laryngologists, ensuring patient-level partitioning. The rest of the data were divided into training and validation sets in a 90\%-10\% ratio for hyperparameter tuning, resulting in 16,222 images for training, 1,806 for validation, and 7,229 for testing (refer to Table~\ref{tab:FAHSYSU_data_distribution}).

\begin{table}[tb]
\centering
\caption{The configurations of SAM2 fine-tuning.}
\label{tab:sam2_training_details}
\scalebox{0.8}{
\begin{tabular}{ll}
\toprule
\textbf{Parameter}         & \textbf{Details} \\ \midrule
Version                    & \texttt{sam2\_hiera\_tiny} \\
Number of Epochs            & 40 \\
Batch Size                 & 96 \\
Warm-up Steps               & 100 \\
Learning Rate Decay Type    & Cosine \\
Weight Decay               & 0.1 \\
Initial Learning Rate       & 2e-5 \\
Automatic Mixed Precision  & Enabled \\
Optimizer                  & AdamW \\
Image Size                 & 1024 $\times$ 1024 \\
Loss Functions             & Dice Loss + Cross-entropy Loss \\
Data Augmentations         & Normalized only \\
Trainable Components       & Image Encoder, Mask Decoder \\
Frozen Components          & Prompt Encoder \\
Prompt Input               & No prompts used as input \\
Mask Prediction            & Single mask predicted for simplicity \\ \bottomrule
\end{tabular}
}
\end{table}

\begin{table}[tb]
\centering
\caption{Configurations of SAM-Swin in training Stage 1 and Stage 2.}
\label{tab:dual_swin_stages}
\scalebox{0.8}{
\begin{tabular}{lll}
\toprule
\textbf{Parameter}         & \textbf{Stage 1}      & \textbf{Stage 2} \\ \midrule
Version                    & \texttt{SwinV2-B}               & \texttt{SwinV2-B} \\
Number of Epochs            & 50                              & 10 \\
Batch Size                 & 256                             & 256 \\
Warm-up Epochs              & 5                               & 2 \\
Learning Rate Decay Type    & Cosine                          & Cosine \\
Weight Decay               & 0.05                            & 1e-8 \\
Initial Learning Rate       & 3e-4                            & 3e-5 \\
Automatic Mixed Precision  & Enabled                     & Enabled \\
Optimizer                  & AdamW                           & AdamW \\
Image Size                 & 256 $\times$ 256                & 256 $\times$ 256 \\
Loss Functions             & Cross-entropy Loss              & Cross-entropy Loss \\
Data Augmentations         & RandAugment                    & RandAugment \\ \bottomrule
\end{tabular}
}
\end{table}

At first, we fine-tuned SAM2 following the fine-tuning strategy outlined in \citep{medsam2}. To adapt SAM2 for the LPC datasets, we empirically applied the fine-tuning configuration shown in Table~\ref{tab:sam2_training_details}. After this process, SAM2 was used to automatically segment the lesions. In cases where no lesions were detected (i.e. when the predicted mask had zero pixels), a $128 \times 128$ center region was cropped from the entire image. All segmented lesion regions were then used as inputs for the lesion region in subsequent stages. After fine-tuning SAM2, we proceeded to train the SAM2-Swin network, keeping the SAM2's parameters frozen. Our training configuration closely followed the methodologies described in \citep{liu2021swin, liu2022swinv2}, with detailed parameters outlined in Table~\ref{tab:dual_swin_stages}. In Stage 1, we fine-tuned the SAM2-Swin network using SwinV2-B as backbones, utilizing pre-trained weights from ImageNet-1K \citep{deng2009imagenet}. Additionally, we intentionally omitted the CAG loss to prevent potential disruptions, particularly in the shallow layers. In Stage 2, we further refined the last checkpoint for 10 epochs, incorporating CAG loss as an additional class-specific supervision signal, while also decreasing the values for the learning rate, weight decay, and warm-up epochs. Furthermore, to ensure a fair comparison with other baseline models, all input images were resized to $256 \times 256$, and the hyperparameter settings were based on their default configurations \citep{liu2022swinv2, RadFormer, DLGNet, wei2024sam}.

\subsection{Experiment results}
\subsubsection{Baselines}
To comprehensively demonstrate the superiority of our proposed SAM-Swin, we evaluate its performance compared with nine state-of-the-art classification methods across three categories on the internal dataset (FAHSYSU) and external datasets (SAHSYSU and NHSMU). The comparison includes \textbf{CNN-based methods}: VGGNet \citep{VGGNet}, ResNet \citep{resnet}, DenseNet \citep{DenseNet}, and EfficientNet \citep{Efficientnet}; \textbf{Transformer-based methods}: Vision Transformer (ViT) \citep{ViT}, and Swin Transformer V2 (SwinV2) \citep{liu2022swinv2}; \textbf{Dual-branch methods}: RadFormer \citep{RadFormer}, DLGNet \citep{DLGNet}, and SAM-FNet \citep{wei2024sam}.
\begin{itemize}
    \item \textbf{VGGNet}: A CNN model known for its use of small convolutional filters to capture spatial hierarchies.
    \item \textbf{ResNet}: A CNN that introduces residual learning to solve the vanishing gradient problem, enabling the training of much deeper networks.
    \item \textbf{DenseNet}: A CNN where each layer connects to every other layer, prompting feature reuse and efficient gradient flow.
    \item \textbf{EfficientNet}: A family of CNNs that scales network width, depth, and resolution using a compound scaling method to improve efficiency.
    \item \textbf{ViT}: A pure transformer model that applied self-attention mechanisms to patches of an image, capturing long-range dependencies.
    \item \textbf{SwinV2}: A hierarchical transformer model that processes images with shifted windows for efficient computation and better performance across scales.
    \item \textbf{RadFormer}: A dual-branch network that leverages transformer architecture to fuse global and local features, designed for precise gallbladder cancer detection in ultrasound images.
    \item \textbf{DLGNet}: A dual-branch network that combines global and local features to incorporate contextual lesion information, specifically for colon lesion classification.
    \item \textbf{SAM-FNet}: A dual-branch network that fuses global features and lesion-specific features, with lesion regions localized by SAM, for detecting laryngo-pharyngeal tumors.
\end{itemize}

\subsubsection{Classification performance on internal dataset}
Table \ref{model_performance_combined} compares the performance of various state-of-the-art methods, including VGGNet, ResNet, DenseNet, EfficientNet, ViT, SwintV2, RadFormer, DLGNet, and SAM-FNet, with the proposed SAM-Swin on the internal dataset FAHSYSU. Regarding the overall results, the SAM-Swin achieves the highest performance across all the evaluated metrics. It attains an overall accuracy of 94.52\%, precision of 93.75\%, recall of 91.21\%, and F1-score of 92.30\%. This represents a significant improvement over the other techniques, with the next best performer being SAM-FNet, which has an accuracy of 92.14\%, precision of 89.57\%, recall of 88.68\%, and F1-score of 89.08\%. When examining the recall for different classes, the SAM-Swin again demonstrates superior performance. It achieves a recall of 96.54\% for the normal class, 78.73\% for the benign class, and 98.35\% for the malignant class. This suggests that the SAM-Swin is highly effective in accurately detecting and classifying the various types of cases in the datasets. Compared to the other state-of-the-art methods, the SAM-Swin consistently outperforms them across all the evaluation metrics. For instance, it surpasses the VGGNet, ResNet, DenseNet, and EfficientNet methods by a margin of several percentage points in terms of accuracy, precision, recall, and F1-score, respectively. Similarly, it exhibits superior performance compared to the more recent methods like ViT, SwintV2, RadFormer, DLGNet, and SAM-FNet. These results demonstrate the effectiveness and robustness of the SAM-Swin in the context of the FAHSYSU dataset. The substantial improvements over the state-of-the-art techniques highlight the potential of the proposed SAM-Swin for practical applications in laryngo-pharyngeal tumor detection.

\begin{figure}[tb]
    \centering
    \includegraphics[width=\linewidth]{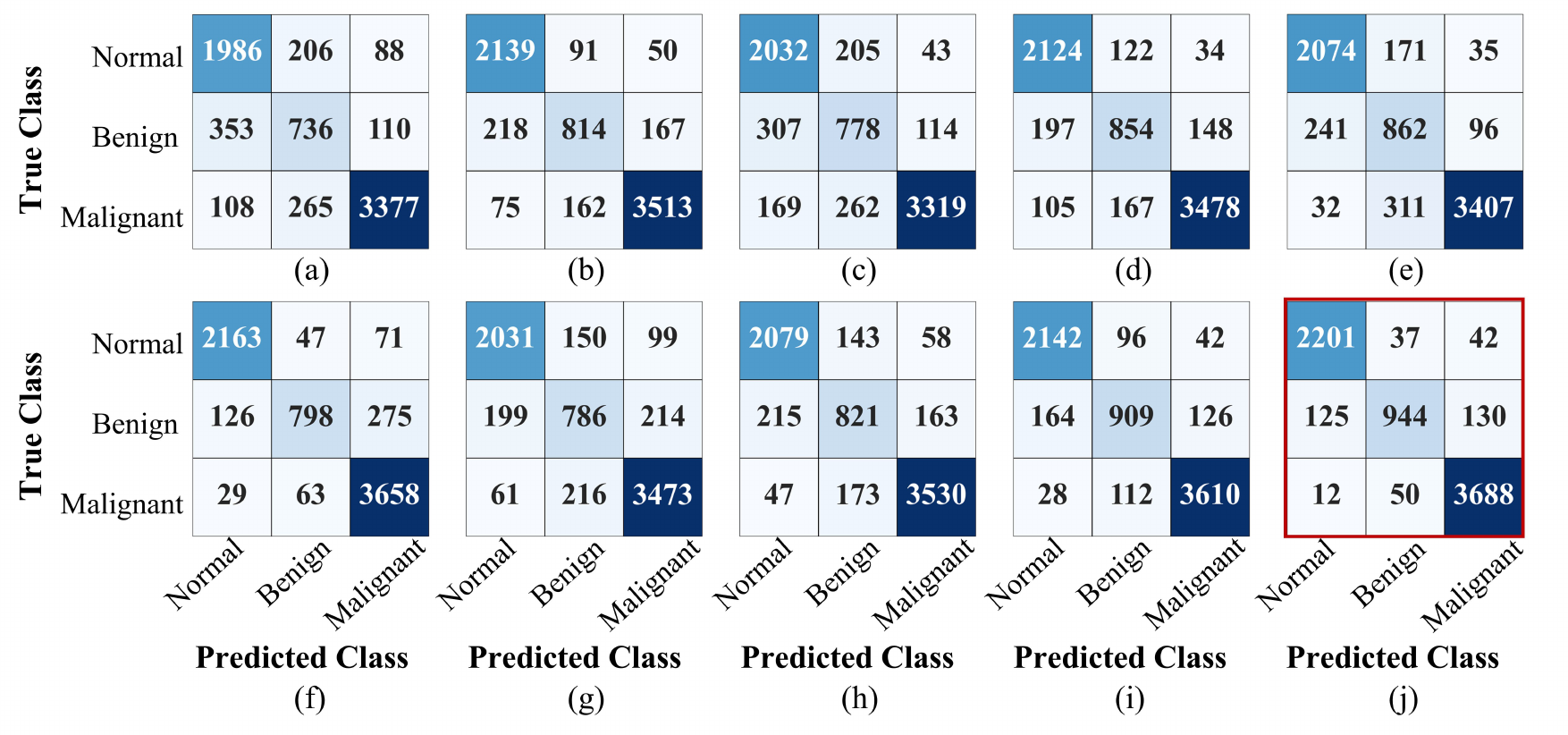}
    \caption{
        Confusion matrices obtained by our proposed SAM-Swin and other comparative methods on the FAHSYSU dataset. (a) VGGNet, (b) ResNet, (c) DenseNet, (d) EfficientNet, (e) ViT, (f) SwinV2, (g) RadFormer, (h) DLGNet, (i) SAM-FNet, (j) SAM-Swin. 
    }
    \label{F.cm}
\end{figure}

Moreover, Fig.~\ref{F.cm} presents the confusion matrices obtained by applying the proposed SAM-Swin method and other comparative methods on the FAHSYSU dataset. The confusion matrices provide a detailed breakdown of the model's performance in classifying the endoscopic images into three categories: normal, benign, and malignant. Starting with the VGGNet model, the confusion matrix (a) shows a significant misclassification between the benign and malignant classes, with a high number of malignant cases being classified as benign. This suggests that the VGGNet model struggles to accurately distinguish between these two clinically relevant classes. The ResNet model (b) exhibits a more balanced performance, with a better separation between the normal, benign, and malignant classes. However, there are still some instances of misclassification, particularly between benign and malignant cases. The DenseNet model (c) demonstrates an improved classification, with clearer boundaries between the three classes. The number of misclassifications between the benign and malignant cases is reduced compared to the previous models. As we move to the more recent models, the confusion matrices show further enhancements in performance. The EfficientNet (d) and ViT (e) models exhibit a higher degree of accuracy in separating the normal, benign, and malignant cases, with a more distinct and compact representation of the class boundaries. The SwintV2 (f), RadFormer (g), and DLGNet (h) models continue to demonstrate improvements, with the confusion matrices displaying sharper and more defined class separations. These models appear to be better equipped to distinguish clinically relevant cases, which is crucial for accurate diagnosis and treatment planning. Finally, the confusion matrices for the SAM-FNet (i) and the proposed SAM-Swin (j) methods show the most distinct and well-defined class boundaries among all the evaluated models. The SAM-Swin method, in particular, exhibits an exceptionally clear separation between the normal, benign, and malignant cases, indicating its superior ability to accurately classify the different types of cases in the FAHSYSU dataset. These results align with the quantitative performance metrics reported in the previous Table \ref{model_performance_combined}, where the SAM-Swin method outperformed the other state-of-the-art techniques across various evaluation measures. The detailed confusion matrices provide a visual representation of the model's classification capabilities, further corroborating the effectiveness of the proposed SAM-Swin approach in the context of the FAHSYSU dataset.

\begin{table*}[tb]
\centering
\caption{Experiment results of the SAM-Swin and other baselines on the FAHSYSU, SAHSYSU, and NHSMU datasets (Unit: \%).}
\label{model_performance_combined}
\scalebox{0.9}{
\begin{tabular}{llcccccccccc}
\toprule
\multirow{2}{*}{Dataset} & \multirow{2}{*}{Method} & \multicolumn{4}{c}{Overall results} & \multicolumn{3}{c}{Recall for different classes} \\ \cmidrule(lr){3-6} \cmidrule(lr){7-9} 
& & Accuracy & Precision & Recall & F1-score & Normal & Benign & Malignant \\
\midrule
\multirow{9}{*}{FAHSYSU} 
& VGGNet \citep{VGGNet}            & 84.37 & 78.87 & 79.51 & 79.14 & 87.11 & 61.38 & 90.05 \\
& ResNet \citep{resnet}            & {89.45} & {86.14} & 85.13 & 85.52 & {93.82} & 67.89 & 93.68 \\
& DenseNet \citep{DenseNet}        & 85.88 & 80.93 & 81.27 & 81.03 & 88.51 & 64.05 & 91.25 \\
& EfficientNet \citep{Efficientnet} & 89.31 & 85.76 & {85.71} & {85.69} & 93.16 & 71.23 & 92.75 \\
& ViT \citep{ViT}                  & 87.74 & 82.93 & 84.57 & 83.65 & 90.96 & {71.89} & 90.85 \\
& SwinV2 \citep{liu2022swinv2}   & 91.54 & 90.82 & 86.30 & 88.04 & 94.83 & 66.56 & 97.52 \\
& RadFormer \citep{RadFormer}      & 87.01 & 82.87 & 82.42 & 82.63 & 89.08 & 65.55 & 92.61 \\
& DLGNet \citep{DLGNet}            & 88.95 & 85.04 & 84.60 & 84.80 & 91.18 & 68.47 & 94.13 \\
& SAM-FNet \citep{wei2024sam}      & 92.14 & 89.57 & 88.68 & 89.08 & 93.95 & 75.81 & 96.27 \\
& SAM-Swin (Ours)                & \textbf{94.52} & \textbf{93.75} & \textbf{91.21} & \textbf{92.30} & \textbf{96.54} & \textbf{78.73} & \textbf{98.35} \\
\midrule
\multirow{9}{*}{SAHSYSU} 
& VGGNet \citep{VGGNet}            & 82.42 & 64.42 & 70.71 & 66.98 & 87.15 & 42.15 & 82.83 \\
& ResNet \citep{resnet}            & 91.07 & 80.18 & 82.37 & 81.22 & 94.81 & 67.05 & 85.24 \\
& DenseNet \citep{DenseNet}        & 84.58 & 68.62 & 72.69 & 70.47 & 89.43 & 45.21 & 83.43 \\
& EfficientNet \citep{Efficientnet} & 87.88 & 74.69 & 81.58 & 77.50 & 90.52 & 68.97 & 85.24 \\
& ViT \citep{ViT}                  & 89.67 & 78.14 & 79.95 & 78.68 & 94.03 & 67.82 & 78.01 \\
& SwinV2 \citep{liu2022swinv2}   & 91.75 & 81.05 & 86.02 & 83.35 & 94.08 & 72.41 & 91.57 \\
& RadFormer \citep{RadFormer}      & 86.80 & 71.57 & 78.64 & 74.61 & 90.30 & 63.98 & 81.63 \\
& DLGNet \citep{DLGNet}            & 86.76 & 72.89 & 80.80 & 75.96 & 89.20 & 67.05 & 86.14 \\
& SAM-FNet \citep{wei2024sam}      & \textbf{92.29} & \textbf{82.71} & 84.52 & 83.59 & \textbf{95.58} & 69.73 & 88.25 \\
& SAM-Swin (Ours)                & 91.28 & 80.14 & \textbf{88.39} & \textbf{83.63} & 92.39 & \textbf{79.69} & \textbf{93.07} \\
\midrule
\multirow{9}{*}{NHSMU} 
& VGGNet \citep{VGGNet}            & 85.37 & 61.20 & 72.90 & 64.52 & 90.08 & 33.95 & 94.68 \\
& ResNet \citep{resnet}            & 92.35 & 77.88 & 85.69 & 81.21 & 94.72 & 68.75 & 93.62 \\
& DenseNet \citep{DenseNet}        & 86.83 & 63.98 & 74.20 & 66.67 & 91.57 & 35.81 & 95.21 \\
& EfficientNet \citep{Efficientnet} & 91.37 & 76.93 & 82.21 & 79.34 & 94.52 & 61.15 & 90.96 \\
& ViT \citep{ViT}                  & 92.47 & \textbf{81.38} & 76.20 & 78.40 & \textbf{97.73} & 47.64 & 83.24 \\
& SwinV2 \citep{liu2022swinv2}   & 92.41 & 77.99 & 87.33 & 82.06 & 94.42 & 69.43 & \textbf{98.14} \\
& RadFormer \citep{RadFormer}      & 87.94 & 68.37 & 83.50 & 73.25 & 89.77 & 66.05 & 94.68 \\
& DLGNet \citep{DLGNet}            & 88.63 & 68.83 & 82.65 & 73.95 & 90.87 & 64.53 & 92.55 \\
& SAM-FNet \citep{wei2024sam}      & 92.34 & 77.62 & 85.99 & 81.17 & 94.63 & 68.92 & 94.41 \\
& SAM-Swin (Ours)                & \textbf{93.06} & 80.25 & \textbf{88.20} & \textbf{83.83} & 94.94 & \textbf{71.79} & 97.87 \\
\bottomrule
\multicolumn{9}{l}{$^{\mathrm{1}}$The best performance is in \textbf{bold}.}
\end{tabular}
}
\end{table*}

\subsubsection{Classification performance on external datasets}
As shown in Table \ref{model_performance_combined}, the SAM-Swin shows similar dominance results on external datasets of SAHSYSU and NHSMU as the internal dataset of FAHSYSU. As for the NHSMU dataset, the SAM-Swin achieves an overall accuracy of 93.06\%, which is the highest among the compared baselines. The precision of 80.25\% and recall of 88.20\% contribute to an impressive F1-score of 83.83\%. Regarding the recall for different classes on the NHSMU dataset, the SAM-Swin again exhibits its strengths. It attains a recall of 94.94\% for the normal class, 71.79\% for the benign class, and 97.87\% for the malignant class. This showcases the model's ability to effectively identify and classify the various types of cases in the external NHSMU dataset. Compared to the other methods on the NHSMU dataset, the SAM-Swin outperforms its counterparts across the board. It surpasses the performance of VGGNet, ResNet, DenseNet, EfficientNet, ViT, SwintV2, RadFormer, DLGNet, and SAM-FNet in terms of overall accuracy, precision, recall, and F1-score. The consistent and exceptional performance of the SAM-Swin method on the SAHSYSU and NHSMU datasets, in addition to the previously discussed FAHSYSU dataset, demonstrates the robustness and generalizability of the proposed SAM-Swin. This highlights its potential for practical applications in laryngo-pharyngeal tumor detection, where accurate and reliable classification is crucial.

\subsubsection{Visualization analysis of Grad-CAM}
\begin{figure*}[tb]
    \centering
    \includegraphics[width=\linewidth]{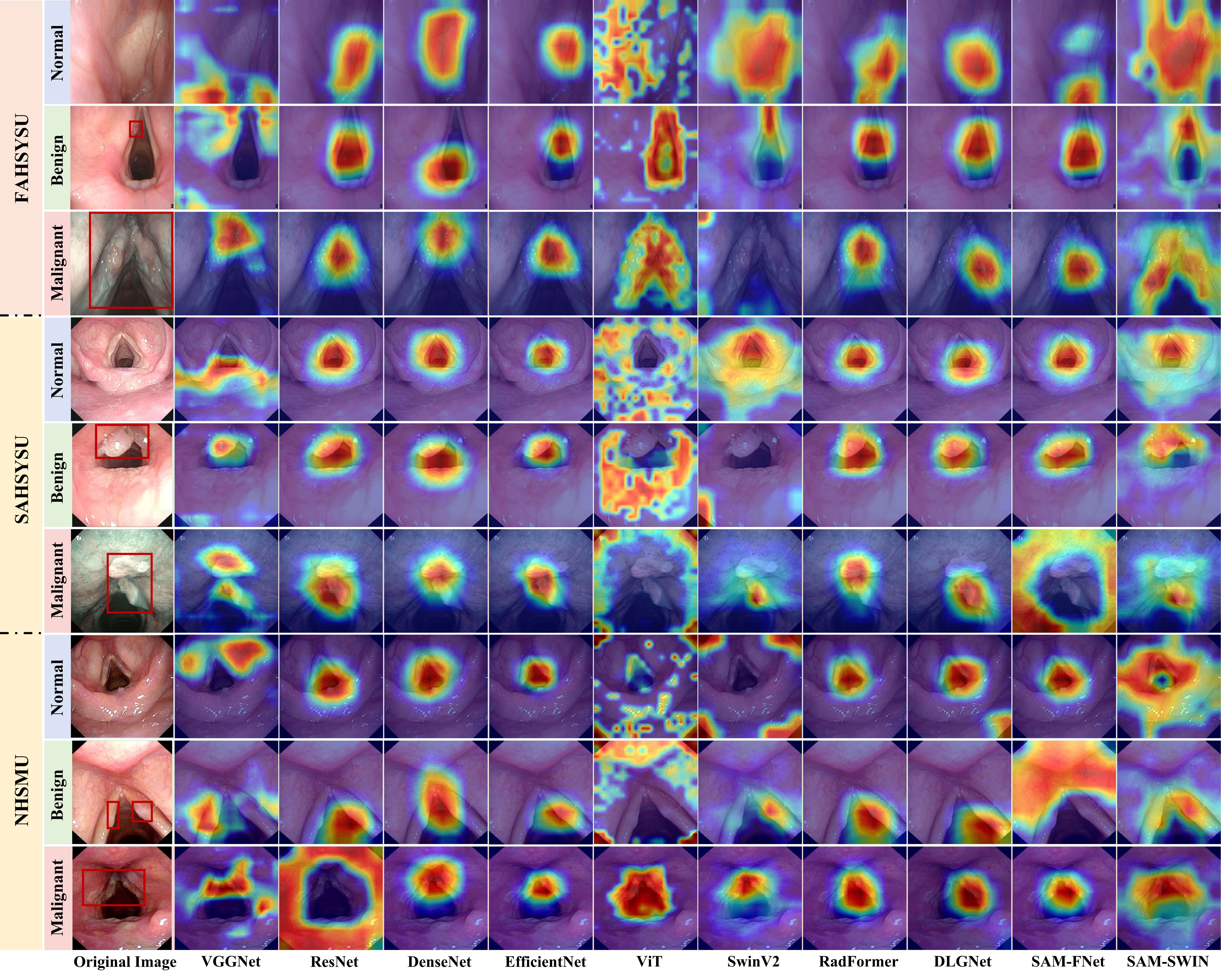}
    \caption{Illustrations of the Grad-CAM visualization on the FAHSYSU, SAHSYSU, and NHSMU datasets.}
    \label{F.cam_vis}
\end{figure*}

The Grad-CAM visualizations shown in Fig.~\ref{F.cam_vis} provide an in-depth analysis for the FAHSYSU, SAHSYSU, and NHSMU datasets. Starting with the FAHSYSU dataset, the Grad-CAM maps reveal some interesting observations. The VGGNet, ResNet, and DenseNet models tend to focus on broader, more diffuse regions across the input images, often failing to precisely capture the key features when the benign tumor is small. Similarly, CNN-based methods like EfficientNet, DLGNet, and SAM-FNet tend to concentrate on limited parts of the tumors, potentially leading to misclassification between tumor types due to the lack of contextual information. Transformer-based architectures such as ViT and SwinV2 benefit from a broader receptive field, allowing for greater attention coverage. However, they often struggle to emphasize lesion regions effectively, lacking the capacity to capture detailed, lesion-specific features. In contrast, the SAM-Swin model hones in on the most discriminative regions within the input images, even for small and superficial tumors. It also captures the complete tumor region, including boundary details and surrounding tissue, enabling it to harness contextual information essential for fine-grained tumor differentiation \citep{WANG2022102395}. A similar pattern emerges for the SAHSYSU and NHSMU datasets. While the other state-of-the-art models show varying degrees of attention to different parts of the images, the SAM-Swin consistently demonstrates a more targeted and comprehensive focus on the most informative visual cues. This precise and focused attention exhibited by the SAM-Swin aligns with its superior performance reported in the previous table. By effectively capturing and utilizing the most relevant features in the input images, the SAM-Swin is able to achieve higher accuracy, precision, recall, and F1-score compared to the other deep learning techniques. The Grad-CAM visualizations offer a visual interpretation of the inner workings of these models, providing insights into their decision-making processes. The clear and focused attention of the SAM-Swin suggests its ability to learn and leverage the most discriminative features in a more efficient and effective manner, contributing to its overall superior performance on the FAHSYSU, SAHSYSU, and NHSMU datasets.

\subsubsection{Ablation experiments}
\begin{table*}[tb]
\centering
\caption{Ablation experiments performance on the FAHSYSU dataset (Unit: \%).}
\label{ablation_study}
\scalebox{0.9}{
\begin{tabular}{cccccccccccc}
\toprule
\multirow{2}{*}{Variant} & \multirow{2}{*}{WIB} & \multirow{2}{*}{LRB} & \multirow{2}{*}{MS-LAEM} & \multirow{2}{*}{CAG loss} & \multicolumn{4}{c}{Overall results} & \multicolumn{3}{c}{Recall for different classes} \\ 
\cmidrule(lr){6-9} \cmidrule(lr){10-12}
& & & & & Accuracy & Precision & Recall & F1-score & Normal & Benign & Malignant \\
\midrule
M1 & \checkmark & & & & 91.54 & 90.82 & 86.30 & 88.04 & 94.83 & 66.56 & 97.52 \\
M2 & & \checkmark & & & 89.78 & 87.70 & 83.45 & 84.87 & 94.08 & 59.38 & 96.88 \\
M3 & \checkmark & \checkmark & & & 93.53 & 92.73 & 89.57 & 90.87 & 96.45 & 74.40 & 97.87 \\
M4 & \checkmark & \checkmark & \checkmark & & 94.05 & 93.11 & 90.52 & 91.63 & 96.19 & 77.23 & 98.13 \\
M5 (SAM-Swin) & \checkmark & \checkmark & \checkmark & \checkmark & \textbf{94.52} & \textbf{93.75} & \textbf{91.21} & \textbf{92.30} & \textbf{96.54} & \textbf{78.73} & \textbf{98.35} \\
\bottomrule
\multicolumn{12}{l}{$^{\mathrm{1}}$The best performance is in \textbf{bold}.}
\end{tabular}
}
\end{table*}

Table~\ref{ablation_study} presents the results of the ablation experiments conducted on the FAHSYSU dataset for the SAM-Swin and its variants. These experiments aim to evaluate the impact of different components and architectural choices on the overall performance of the model. The variants investigated include M1, M2, M3, M4, and the final proposed method M5 (SAM-Swin). Specifically:

\begin{itemize}
    \item \textbf{M1 (WIB)}: This configuration utilizes solely the WIB module. It serves as a baseline for assessing the efficacy of the full image processing capabilities in isolation.
    \item \textbf{M2 (LRB)}: In this model, only the LRB model is employed. This setup allows for an in-depth analysis of the region-based processing mechanisms without the influence of the whole image context.
    \item \textbf{M3 (WIB + LRB)}: This variant integrates both the WIB and the LRB modules. By combining these two modules, we aim to evaluate the synergistic effects of processing both whole images and localized regions.
    \item \textbf{M4 (WIB + LRB + MS-LAEM)}: Building on the M3 configuration, this model incorporates the MS-LEAM module. This integration is hypothesized to further enhance the model's ability to adaptively learn from multi-scale features.
    \item \textbf{M5 (SAM-Swin)}: The final configuration encompasses all previous components: WIB, LRB, MS-LEAM), and incorporates CAG loss. This comprehensive model aims to leverage the strengths of all preceding variants for optimal performance in complex tasks.
\end{itemize}

Table~\ref{ablation_study} summarizes the performance metrics of the various model variants, including accuracy, precision, recall, and F1-score, as well as the recall values for the three classes: normal, benign, and malignant. The M1 variant achieves a commendable overall accuracy of 91.54\%, with a precision of 90.82\%, recall of 86.30\%, and an F1-score of 88.04\%. The M2 variant achieves an overall accuracy of 89.78\%, with a precision of 87.70\%, a recall of 83.45\%, and an F1-score of 84.87\%. It can be observed that the M2 variant experiences a decrease across all metrics, suggesting that the absence of comprehensive contextual information may hinder its ability to distinguish lesions effectively, leading to a decline in overall performance. As we progress through the ablation study with subsequent variants (M3 and M4), we observe incremental enhancements in performance metrics. Notably, the M4 variant demonstrates significant improvements, achieving an overall accuracy of 94.05\%, precision of 93.11\%, recall of 90.52\%, and an F1-score of 91.63\%. The final model, SAM-Swin (M5), showcases the highest performance metrics, with an accuracy of 94.52\%, precision of 93.75\%, recall of 91.21\%, and an F1-score of 92.30\%. The recall for the normal, benign, and malignant classes is particularly noteworthy, at 96.54\%, 78.73\%, and 98.35\%, respectively. The consistent improvements observed across the different variants, especially in the recall for benign and malignant classes, underscore the effectiveness of the architectural choices and components integrated within the SAM-Swin model. 

\begin{figure}[tb]
    \centering
    \includegraphics[width=\linewidth]{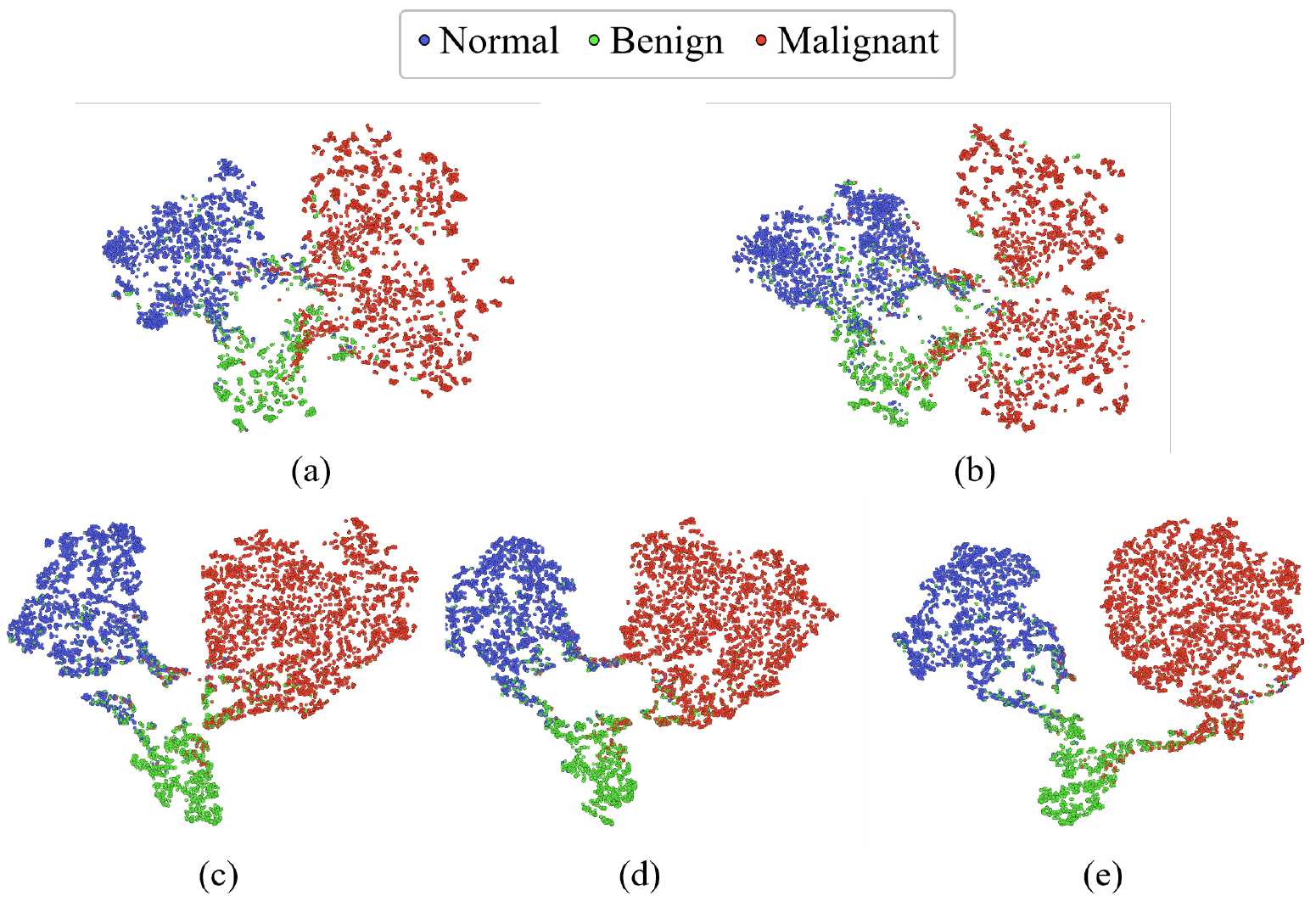}
    \caption{
        The t-SNE visualization of the ablation experiment on the FAHSYSU dataset. Among them, (a) M1 (WIB only), (b) M2 (LRB only), (c) M3 (WIB+LRB), (d) M4 (WIB+LRB+MS-LAEM), (e) M5 (SAM-Swin, all components).
    }
    \label{F.t-sne}
\end{figure}

Furthermore, Fig.~\ref{F.t-sne} illustrates the t-SNE visualizations derived from the ablation experiment results on the FAHSYSU dataset. The t-SNE technique effectively projects high-dimensional feature representations onto a two-dimensional space, facilitating a visual exploration of cluster structures and class separability across different model variants: M1, M2, M3, M4, and the final SAM-Swin approach (M5). In the t-SNE plot for the M1 variant (a) and the M2 variant (b), a degree of class separation is evident; however, notable overlap persist, particularly with the benign class. This suggests that further refinement is necessary for improved discrimination among case types. As the ablation progresses, the M3 variant (c) exhibits a more pronounced separation between classes, with clearer distinctions for normal and malignant samples, as well as improved differentiation of benign samples. The M4 variant (d), which incorporates additional architectural components, further enhances class separability, resulting in more distinct and tighter clusters for each class. The t-SNE visualization of the SAM-Swin method (M5) (e) achieves the most remarkable class separation among all variants, with minimal overlap between normal, benign, and malignant samples. This clear delineation of class boundaries indicates that the SAM-Swin approach has effectively learned highly discriminative features.

\subsection{Analysis of SAM2-GLLM}

\begin{table}[tb]
\centering
\caption{Average IoU and Dice scores of different segmentation methods on the FAHSYSU (Unit: \%).}
\label{tab:iou_dice}
\begin{tabular}{lcc}
\toprule
\textbf{Model} & \textbf{IoU (Average)} & \textbf{Dice (Average)} \\
\midrule
UNet & 47.27 & 59.22 \\
Mask-RCNN & 60.34 & 69.97 \\
SAM  & 60.96 & 72.07 \\
SAM2 & \textbf{63.63} & \textbf{74.55} \\
\bottomrule
\end{tabular}
\end{table}

As shown in Table~\ref{tab:iou_dice}, SAM2 achieves the highest segmentation performance among the segmentation methods, with an average IoU of 63.63\% and an average Dice score of 74.55\%. Compared to traditional models like UNet (IoU: 47.27\%, Dice: 59.22\%) and Mask-RCNN (IoU: 60.34\%, Dice: 69.97\%), SAM2 demonstrates significant improvements. Furthermore, it surpasses the SAM by 2.67\% in IoU and 2.48\% in Dice, highlighting its enhanced capability for accurate lesion localization. These results validate the effectiveness of the SAM2-GLLM in improving segmentation precision in the laryngo-pharyngeal tumor context.

\begin{figure}[tb]
    \centering
    \includegraphics[width=\linewidth]{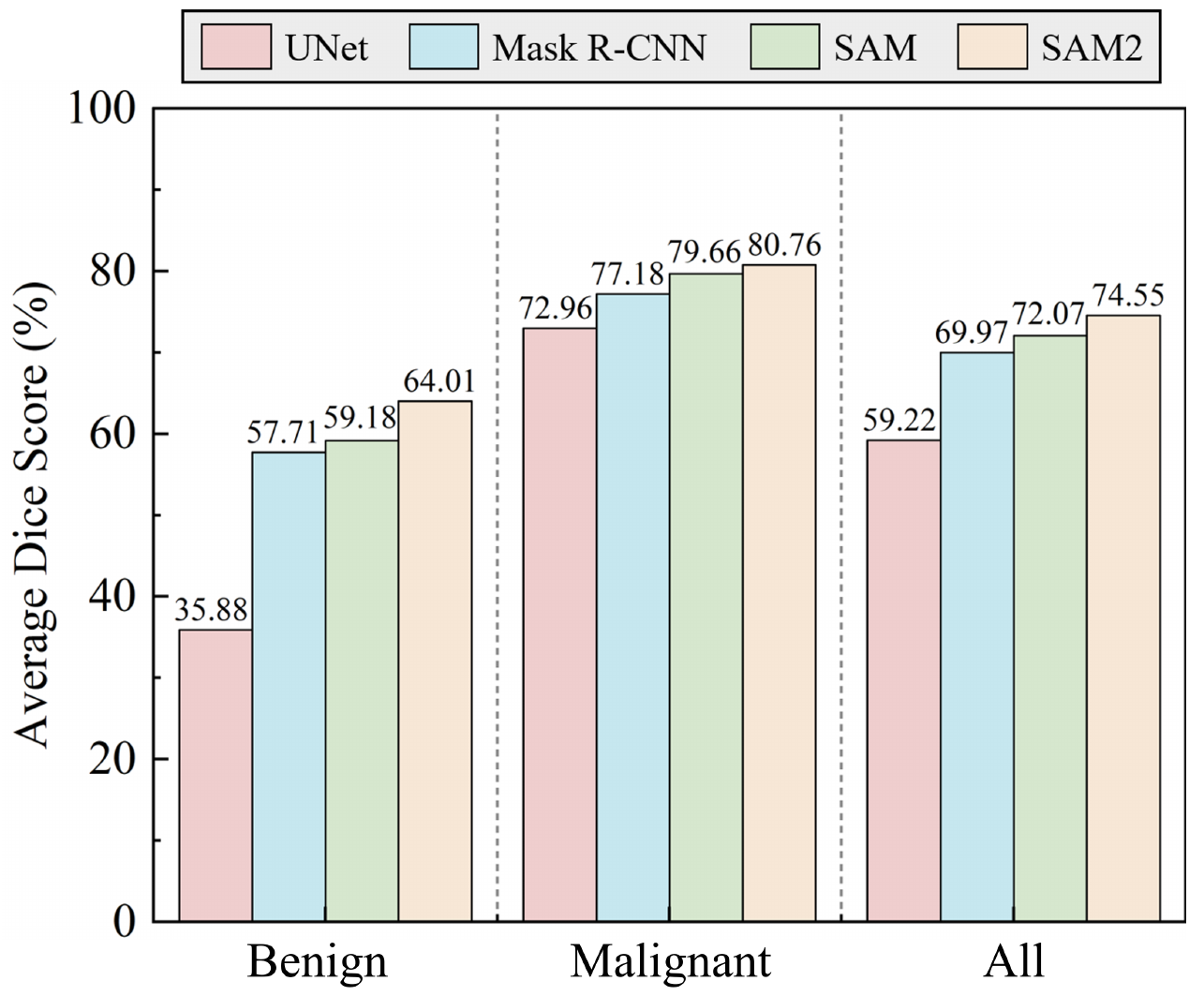}
    \caption{
        Average Dice scores of different segmentation methods across benign, malignant, and combined (All) categories on the FAHSYSU dataset.
    }
    \label{F.dice}
\end{figure}

As illustrated in Fig.~\ref{F.dice}, SAM2 demonstrates superior segmentation performance across different tumor categories. Notably, SAM2 achieves a substantial improvement in segmenting benign lesions, achieving a Dice score of 64.01\% compared to 35.88\% for UNet, 57.71\% for Mask R-CNN, and 59.18\% for SAM. Additionally, SAM2 attains an average Dice score of 80.76\%, surpassing other methods such as UNet (72.96\%), Mask R-CNN (77.18\%), and SAM (79.66\%). This significant advantage in lesion segmentation highlights SAM2’s enhanced capability to accurately locate less-defined lesion boundaries, underscoring its potential as a reliable lesion localization module in challenging cases.

\subsection{Analysis of MS-LAEM}
\begin{figure}[tb]
    \centering
    \includegraphics[width=\linewidth]{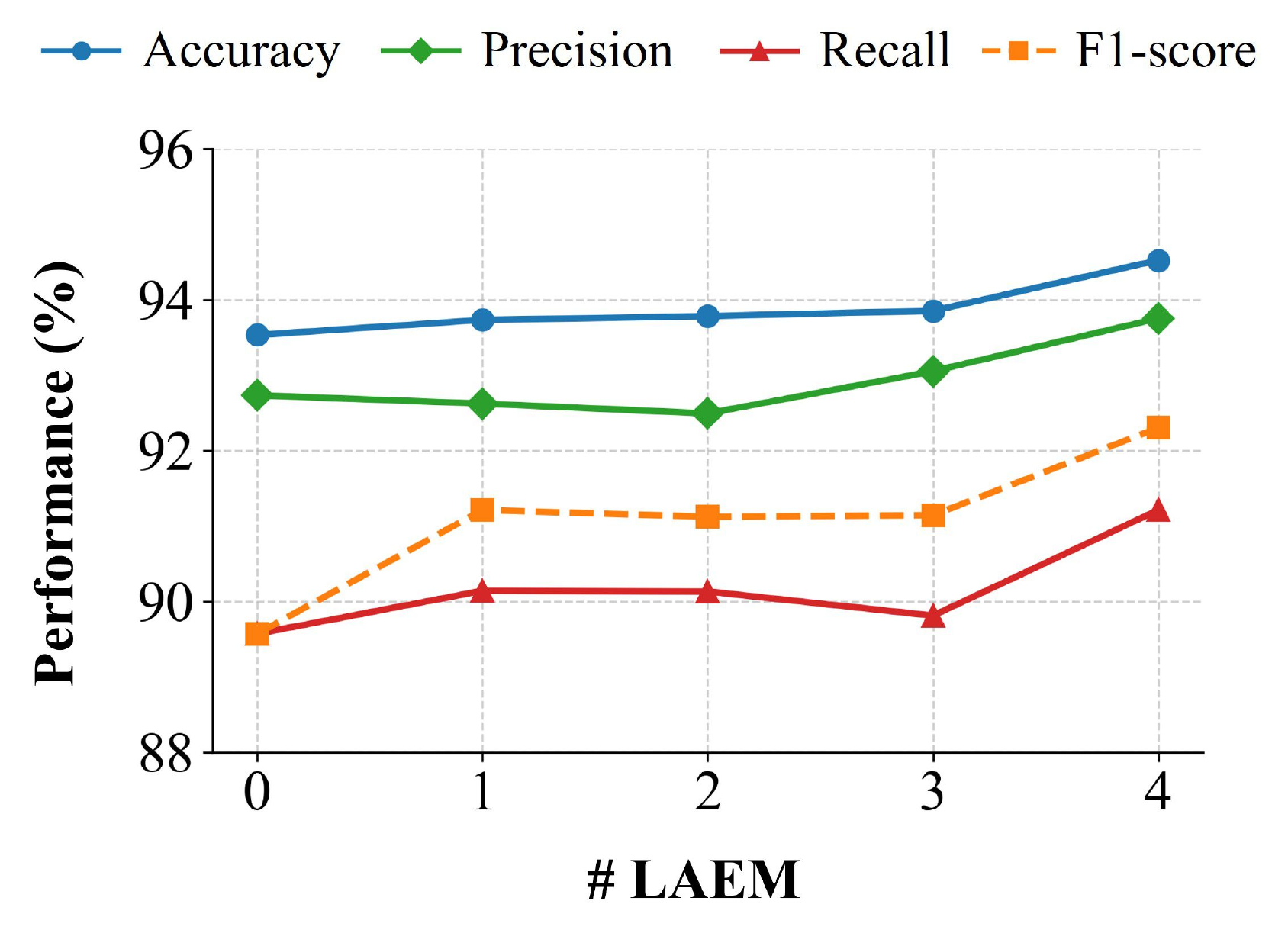}
    \caption{
        Results of the proposed SAM-Swin with different numbers of lesion-aware enhancement modules on the FAHSYSU dataset.
    }
    \label{F.num_layers}
\end{figure}

\textbf{Influence of the number of LAEMs.} As mentioned in Section~\ref{sec:MS-ALEM}, the MS-LAEM contains several LAEMs applied to different stages. To evaluate their impact on classification performance, we varied the number of inserted LAEMs from 0 to 4. A setting of "0" represents no enhancement, while "1" to "4" indicates the incremental inclusion of LAEMs from later to earlier stages, with "4" utilizing all stages, and the results are shown in Fig.~\ref{F.num_layers}. It is observed that from 0 to 1 LAEM, there is a noticeable improvement in both recall and F1-score, demonstrating the effectiveness of introducing the LAEM for feature enhancement. Additionally, from 1 to 3 LAEMs, accuracy steadily increases, suggesting that incorporating additional LAEMs at multiple stages continues to refine the model's performance. At 4 LAEMs, the best results are achieved across all metrics, indicating that fully leveraging multi-scale features provides the most comprehensive enhancement.

\begin{figure}[tb]
    \centering
    \includegraphics[width=\linewidth]{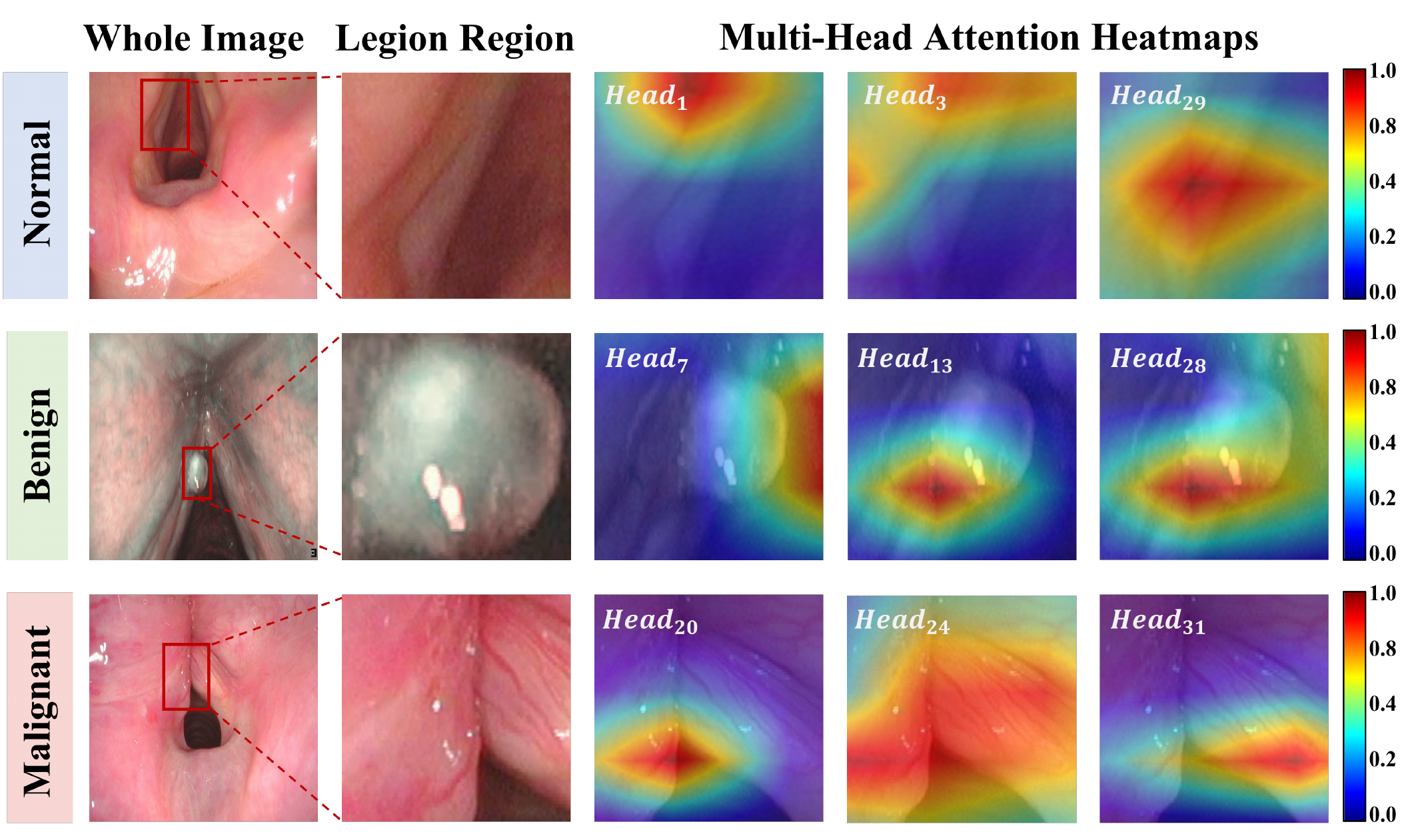}
    \caption{
        Visualizations of the multi-head lesion-aware attention maps at Stage 4, illustrating how different heads focus on distinct parts of the lesion region within the whole image.
    }
    \label{F.att_maps}
\end{figure}

\textbf{Visualization of lesion-aware attention maps.} To verify the effectiveness of LAEM, we plotted the multi-head attention heatmaps from stage four, as illustrated in Fig.~\ref{F.att_maps}. The multi-head attention mechanism allows each head to focus on different areas within the lesion region, enabling the model to capture diverse contextual information surrounding the lesion, which enhances its ability to discriminate between lesion types.

\subsection{Hyperparameters tuning} \label{sec:hyper}
\begin{figure}[tb]
    \centering
    \includegraphics[width=\linewidth]{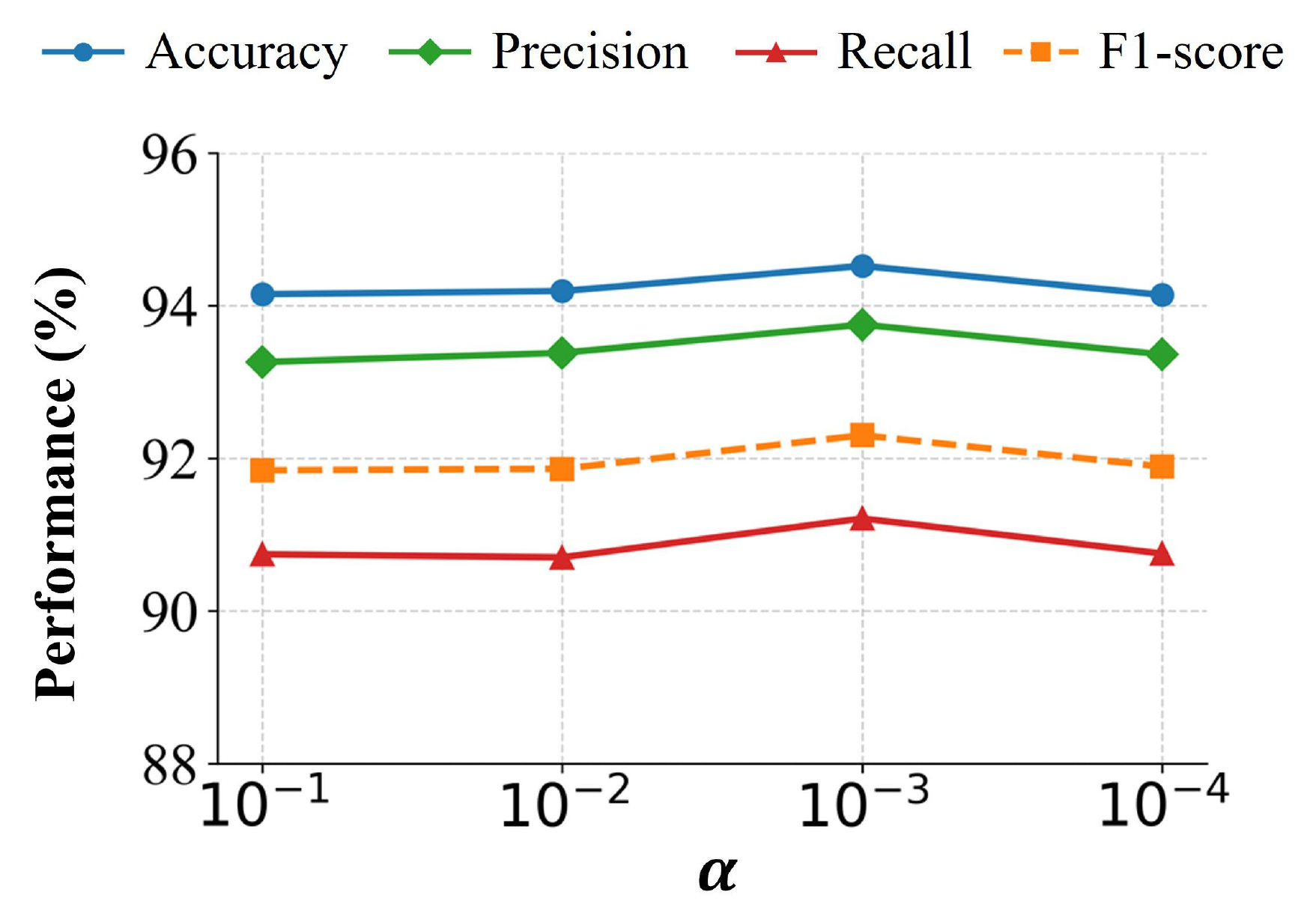}
    \caption{
        Results of the SAM-Swin with different values of $\alpha$ for the class-aware guidance loss on the FAHSYSU dataset.
    }
    \label{F.hyper}
\end{figure}

To analyze the impact of different weights for the multi-scale CAG loss as discussed in Section~\ref{sec:cag_loss} on classification performance, we varied the $\alpha$ over a range of $\{ 10^{-1}, 10^{-2}, 10^{-3}, 10^{-4} \}$, as depicted in Fig.~\ref{F.hyper}. It is clear that the best performance is achieved with a weight of $10^{-3}$. This suggests that the weights like $10^{-1}$ and $10^{-2}$ are too large, causing the model to over-focus on class-specific features, while $10^{-4}$ is too small to provide sufficient guidance. The weight $10^{-3}$ strikes an optimal balance, allowing the model to leverage class-specific supervision effectively, leading to the highest overall performance.

\section{Conclusion}
\label{sec conclusion}
In this study, we propose a novel SAM-driven Dual-Swin Transformer framework integrated with adaptive lesion enhancement, tailored for the detection of laryngo-pharyngeal tumors. It mainly consists of four key components: SAM2-GLLM, WIB, LRB, and MS-LAEM. Furthermore, we implement a multi-scale CAG loss function, which provides class-specific supervision to improve the model's ability to differentiate between various tumor categories. Comprehensive experiments conducted on three distinct datasets, including both internal and external sources, demonstrate the effectiveness of our SAM-Swin model, achieving overall accuracies of 94.52\%, 91.28\%, and 93.06\% in terms of FAHSYSU, SAHSYSU, and NHSMU, respectively. These results significantly surpass existing state-of-the-art approaches in laryngo-pharyngeal tumor detection. In addition, Grad-CAM visualizations, ablation studies, and detailed module effectiveness analyses further validate the reliability and superiority of our proposed method. Our comprehensive analysis underscores the potential of the SAM-Swin framework as a powerful tool for LPC detection, paving the way for more effective diagnostic applications in clinical practice.

\section*{Acknowledgments}
We note that a regular conference version of this paper appeared in the 2024 IEEE International Conference on Bioinformatics and Biomedicine (IEEE BIBM 2024) \citep{wei2024sam}. This work is partially supported by the National Natural Science Foundation of China (62473267), the Basic and Applied Basic Research Project of Guangdong Province (2022B1515130009), the Special subject on Agriculture and Social Development, Key Research and Development Plan in Guangzhou (2023B03J0172), and the Natural Science Foundation of Top Talent of SZTU (GDRC202318).




\end{document}